%% file: main.tex
\documentclass[dvipsnames,format=sigconf,anonymous=false,review=false,table]{acmart}

\usepackage{amsmath}
\usepackage{enumitem}
\usepackage{soul}

\copyrightyear{2023} 
\acmYear{2023} 
\setcopyright{acmlicensed}\acmConference[GECCO '23]{Genetic and Evolutionary Computation Conference}{July 15--19, 2023}{Lisbon, Portugal}
\acmBooktitle{Genetic and Evolutionary Computation Conference (GECCO '23), July 15--19, 2023, Lisbon, Portugal}
\acmPrice{15.00}
\acmDOI{10.1145/3583131.3590502}
\acmISBN{979-8-4007-0119-1/23/07}

\begin{document}

\title{Solving Novel Program Synthesis Problems with Genetic Programming using Parametric Polymorphism}

\author{Edward Pantridge}
\orcid{0000-0003-0535-5268}
\affiliation{
  \institution{Swoop}
  \city{Cambridge} 
  \state{Massachusetts} 
  \country{USA}
}
\email{ed@swoop.com}

\author{Thomas Helmuth}
\orcid{0000-0002-2330-6809}
\affiliation{
	\institution{Hamilton College}
	\city{Clinton}
	\state{New York}
	\country{USA}
}
\email{thelmuth@hamilton.edu}


\begin{abstract}
Contemporary genetic programming (GP) systems for general program synthesis have been primarily concerned with evolving programs that can manipulate values from a standard set of primitive data types and simple indexed data structures. In contrast, human programmers do not limit themselves to a small finite set of data types and use polymorphism to express an unbounded number of types including nested data structures, product types, and generic functions. Code-building Genetic Programming (CBGP) is a recently introduced method that compiles type-safe programs from linear genomes using stack-based compilation and a formal type system. Although prior work with CBGP has shown initial demonstrations of polymorphism inside evolved programs, we have provided a deeper exploration of these capabilities through the evolution of programs which make use of generic data types such as key-value maps, tuples, and sets, as well as higher order functions and functions with polymorphic type signatures. In our experiments, CBGP is able to solve problems with all of these properties, where every other GP system that we know of has restrictions that make it unable to even consider problems with these properties. This demonstration provides a significant step towards fully aligning the expressiveness of GP to real world programming.
\end{abstract}

%
%



\maketitle

\section{Introduction}

Genetic programming has recently turned more of its focus on the task of general program synthesis~\cite{oneill:2019:AutomaticProgrammingOpen}.
Fitting within the \textit{programming by example} framework~\cite{gulwani:2011:flashfill},
the goal of the task is to automatically produce a program specified by a set of training examples showing correct input/output behavior for the desired program.
Programming by example systems are not given a natural language description of the program like with large language model synthesis systems~\cite{chen_evaluating_2021, austin_program_2021, Sobania:2022:GECCO}, nor a sketch~\cite{solar2013program, Bladek:2017:EuroGP} or underlying grammar for the program~\cite{Alur:2013:SyGuS, alur:2018:CACM} like in syntax-guided synthesis. This work only considers programming by example techniques that take specifications in the form of training cases.


Genetic programming systems which manipulate multiple data types and perform basic iteration and control flow have existed for decades~\cite{Spector:2005:push3}. These systems vary in exactly which types and operations are supported, but generally include the primitive data types $Boolean$, $Int$, $Float$, $String$, and $Character$. GP systems also commonly support indexed data structures, such as lists or vectors, that contain these primitive types. Furthermore, GP systems for program synthesis must be able to produce some form of conditional control flow and iteration or recursion. 

In contrast, human programmers use a vast set of computational paradigms and language features to create programs that solve ``real world'' problems. This includes writing programs that introduce new abstractions, create new data types through composition, and define new functions. Difficult problems are solved by the non-trivial composition of simple components, therefore a finite set of simple non-composable types and a fixed function set is insufficient in most problem domains.

With this in mind, it seems obvious that the expressive power of most contemporary GP systems is woefully weak compared to the aspirations of our field. This paper takes a step in the right direction by expanding the use of polymorphism in a GP to allow for arbitrary nesting of structures, generic transformation of data, and functions with polymorphic type signatures. In order to properly demonstrate the expanded landscape of problems GP can be applied to, we also present a suite of synthetic benchmarks which require the use of new data types and behaviors. To our knowledge, no existing GP system can even be applied to all of these problems, let alone solve them.

In the next section we describe Code-building GP (CBGP), the GP method that we enhanced for this work, which can solve most of our new benchmark problems. In the subsequent section we provide a precise definition of  ``polymorphism'' for the purposes of this paper. We then summarise prior work if GP methods that handle polymorphism. The remaining sections describe our benchmark problems, experimental methods, and results respectively. Finally we conclude with a discussion of some insights from our experimental results and directions for future work.

\section{Code-building Genetic Programming}
\label{sec:CBGP}

CBGP evolves general programs using linear genomes and a stack-based compilation process that compiles genomes into type-safe abstract syntax trees (ASTs). Although originally introduced in~\cite{Pantridge:2020:GECCO:CBGP}, the introduction of a formal type system in~\cite{Pantridge:2022:GECCO:fCBGP} allowed for a more rigorous definition of CBGP's capabilities, such as the use of polymorphic functions and control flow via higher order functions.

Genomes in the CBGP system are variable length sequences of ``genes'' which are nodes found in program ASTs. These include literals, variables, function applications, function abstractions, and \verb|let| local variable bindings. To compile the genome, each gene is processed in order. Genes which correspond to leaf nodes of an AST (such as literals and variables) are annotated with their type and pushed to a stack of ASTs. Genes which correspond to internal nodes of an AST (such as function applications, abstraction, and \verb|let| bindings) will search the AST stack for ASTs with compatible types to create a new composite AST. For example, the function application gene will first cause compilation to search the stack for an AST with a function type and then, depending on the arity of the function and its argument types, compilation will search the stack for additional ASTs to use as inputs to the function. If found, all used ASTs will be removed from the stack and a new composite AST which calls the function on the arguments will be pushed. This stack-based approach allows for genes with unsatisfied constraints to NOOP gracefully. After the entire genome has been compiled, an AST for the problem's output type is selected from the ASTs on the stack and returned as the genomes corresponding program. The compilation process also includes a mechanism for dynamically resolving local variable references depending on which variables are in scope at different locations of the program.

Evolution is driven by a standard generational genetic algorithm starting from a population of randomly generated genome sequences, where genes are sampled from a weighted \textit{genetic source} (the set of literals, variables, and other genes that can appear in CBGP genomes)~\cite{Helmuth:2020:ALife:source}. Each generation, genomes are compiled into ASTs which are loaded into the host language's runtime as native functions. These functions are evaluated on a dataset of training cases from which a vector of error values is produced to inform parent selection. Variation of parent genomes (via mutation and/or crossover) is performed to create the next generation.

CBGP has demonstrated trade-offs compared to other contemporary GP methods. The generalization of evolved programs on unseen data is higher for CBGP than other GP systems, however it fails to solve problems which require non-trivial control flow~\cite{Pantridge:2022:GECCO:fCBGP}. Additionally, the creation and execution of native functions dramatically reduces the execution costs compared to GP systems that incur the overhead of a custom program execution model.

This paper presents results using a functional CBGP system identical to that of~\cite{Pantridge:2022:GECCO:fCBGP}\footnote{The implementation CBGP system can be found here: \url{https://github.com/erp12/cbgp-lite/releases/tag/GECCO-2023}.}. We configure the system with an enhanced set of functions which operate on a larger, more generic, set of data types, and apply the system to novel benchmark problems that we believe would be unapproachable using other contemporary GP systems.

\section{Parametric Polymorphism}


When multiple data types share a common interface, we call that interface \textit{polymorphic}; when an interface only supports a single type, it is \textit{monomorphic}. Polymorphism has been a core feature of most popular programming languages since the popularization of Algol68 and ML in 1970s~\cite{Mailloux:1969:Algol68, Gordon:2000:LfcToHolShortHistory}. There are many forms of polymorphism, including ad-hoc polymorphism, parametric polymorphism, subtyping, and row polymorphism~\cite{Cardelli:1985:TypesAbstractionPolymorph}. The remainder of this paper will use ``polymorphism'' to refer specifically to parametric polymorphism unless otherwise specified.

\textit{Parametric polymorphism} refers to the use of generic data types which can produce or consume values of any type. The definition of a polymorphic type includes one or more ``type variables'' which get bound to a concrete type at the call site where the polymorphic type is used.

The most common example of parametric polymorphism is a collection (aka data structure). The job of a collection is to hold some number of items and provide an interface for accessing them. This typically does not require knowledge of the items' data type(s) which allows us to implement the collection's behavior generically such that each instance of the collection can use an arbitrary item type. In this paper, we only consider ``typed'' collections where all items must belong to the same type, but we acknowledge that some type systems support forms of polymorphism that allow for heterogeneous collections. 

The Hindley-Milner (HM) type system is one of the earliest examples of a type system which supports parametric polymorphism and provides type checking and type inference capabilities~\cite{Hindley:1969:HMTypeSystem}. An implementation of the HM type system, such as Algorithm W, can analyze the abstract syntax tree (AST) of a purely functional program and determine the most general type of the values produced by every expression~\cite{Milner:1978:HMTypeSystem}. If no such type can be found, the program is proved to be not type safe. This analysis does not require executing the program.

\subsection{Type Constructors}
\label{sec:type-constructors}

One place where polymorphism arises in Hindley-Milner based systems is via type constructors. This is a common way of implementing interfaces for generic data structures and composite types. A type constructor defines a way of building types from other types. For example, the $Vector$ type constructor must be given the data type of a list's elements to produce a concrete list type such as $Vector[Int]$ or $Vector[String]$. The resulting types are considered concrete and can be given to other type constructors. For example, a matrix type could be modeled as $Vector[Vector[Float]]$.

Type constructors can be defined to require multiple types. Examples include key-value structures like $Map[\_, \_]$ or product types such as a 2-element $Tuple[\_, \_]$. Perhaps the most important type constructor in the HM system (and all lambda calculus based systems) is the function type constructor.  A function type is defined by one or more argument types and a return type, denoted as $(A_1, ..., A_n) \rightarrow R$ for a function of arity $n$.

The CBGP system used for the experiments presented in this paper is supported by a HM type system which includes the following type constructors.
\begin{itemize}
    \item $Vector[\_]$
    \item $Set[\_]$
    \item $Map[\_, \_]$
    \item $Tuple[\_, \_]$
    \item Unary Function: $(\_) \rightarrow \_$
    \item Binary Function: $(\_, \_) \rightarrow \_$
    \item Trinary Functions: $(\_, \_, \_) \rightarrow \_$
\end{itemize}

Each of these type constructors has a corresponding a set of polymorphic functions which define the interface of the type.\footnote{Technically, the keys of $Map$ and $Set$ must be comparable with equality, and thus this constraint is an example of \textit{ad-hoc polymorphism}, not parametric polymorphism. In practice, we omit these constraints since our host language Clojure allows for equality on any two values.} We also include functions for converting between different collection types where appropriate. For example, a \verb|map-to-vector| function would accept a value of type $Map[\alpha, \beta]$ and return a value of type $Vector[Tuple[\alpha, \beta]]$ and vice versa for the \verb|vector-to-map| function.

It is common for human programmers to define new type constructors as part of their application, but programs of this kind are outside of the scope of this work.

\subsection{Type Schemes}

Some types produced by type constructors are naturally expressed as abstract types which contain a type variable. These variables indicate locations within the type's structure where any type can be substituted. Typically these variables appear in function types to indicate the function can manipulate values of any type. In the HM type system these abstract types are referred to as ``type schemes.''

The canonical type scheme is the type of the \verb|identity| function. It accepts a single argument of an arbitrary type and returns the same value (of the same type). We denote the \verb|identity| function's type as:
\[
\verb|identity|: \forall \alpha. \alpha \rightarrow \alpha
\]
which can be read as: ``For all possible types $\alpha$, a function which takes an instance of $\alpha$ and returns an instance of $\alpha$.''

To demonstrate a more complex type scheme, consider the type of the \verb|get| function which accesses the value in a $Map$ under a specific key.
\[
\verb|get|: \forall \alpha, \beta. (Map[\alpha, \beta], \alpha) \rightarrow \beta
\]
Notice that this scheme has 2 type variables that each occur in 2 locations within the type. When the \verb|get| function is passed some arguments, all instances of each type variable must be bound to the same type for the composite AST to be type-safe. 

\subsection{Functional Programming Constructs}

Although they are not forms of polymorphism, HM based systems often make heavy use of functional programming constructs such as higher order functions, function composition, and partial function applications. 

Higher order functions (HOF) are functions which take other functions as arguments or return function objects. Commonly used HOF include the collection processing functions \verb|map|, \verb|filter|, \verb|reduce|, and \verb|fold|. In addition to being higher order, these functions are polymorphic:
\begin{align*}
map&: \forall \alpha, \beta . ((\alpha \rightarrow \beta), Vector[\alpha]) \rightarrow Vector[\beta] \\
filter&: \forall \alpha . ((\alpha \rightarrow Boolean), Vector[\alpha]) \rightarrow Vector[\alpha] \\
reduce&: \forall \alpha . (((\alpha, \alpha) \rightarrow \alpha), Vector[\alpha]) \rightarrow \alpha \\
fold&: \forall \alpha, \beta . (((\beta, \alpha) \rightarrow \beta), \beta, Vector[\alpha]) \rightarrow \beta
\end{align*}

Some GP systems, including CBGP, have demonstrated the capability of calling HOF inside evolved programs~\cite{Garrow:2022:GECCO:ECADA:WhyFPSynthesisMatters, Pantridge:2022:GECCO:fCBGP}. This paper extends this capability by providing versions of each HOF for all the collection type constructors mentioned in~\ref{sec:type-constructors} ($Vector$, $Set$, and $Map$). In addition we present the novel capability of being able to evolve programs which themselves are higher order functions.

Function composition is the process of creating a new function by chaining multiple other functions. We denote function composition using the $\circ$ operator. In the following example, \verb|h| is defined as the composition of \verb|f| and \verb|g|. We also show the type of \verb|h| assuming given types for \verb|f| and \verb|g|. 
\begin{align*}
    h &= f \circ g = \lambda x . g(f(x)) \\
    f &: String \rightarrow Int \\
    g &: Int \rightarrow Vector[Int] \\
    h &: String \rightarrow Vector[Int]
\end{align*}
The composition operator is considered a polymorphic function with the following type:
\begin{align*}
    \forall \alpha, \beta, \gamma . (\alpha \rightarrow \beta, \beta \rightarrow \gamma) \rightarrow (\alpha \rightarrow \gamma)
\end{align*}

Partial function application is the process of creating a new function by binding some, but not all, of a function's arguments to fixed values. The result is a new function that only requires inputs for the unbound arguments and will return the result of the original function when called. Below we give the type of the partial application function, \verb|P|, capable of binding 1 argument of a binary function, as well as an example usage on the binary function, \verb|f|, and the literal value 10 to create a new function, \verb|g|. 
\begin{align*}
    P &: \forall \alpha, \beta, \gamma . ((\alpha, \beta) \rightarrow \gamma, \alpha) \rightarrow (\beta \rightarrow \gamma) \\
    f &: (Int, Boolean) \rightarrow String \\
    g &= P(f, 10) \\
    g &: Boolean \rightarrow String 
\end{align*}

The results presented in this paper incorporate function composition and partial application operators to the function set used by evolution. This provides novel expressive power for creating new functions, including polymorphic functions, within the logic of evolved programs.

\section{Approaches to Polymorphism in GP}
\label{sec:polymorphism-gp}

Many modern genetic programming systems are designed to solve general program synthesis problems in which the desired synthesized programs are expected to resemble those which humans write. This implies the use of many data types and language constructs, including various forms of polymorphism.

Most contemporary genetic programming systems do not explicitly support polymorphic types but instead utilize strategies for eliminating polymorphsim while still supporting a wide enough range of data types to be compatible with common benchmark problems.

\subsection{Monomorphization}

Monomorphization is the process of generating a finite set of monomorphic types derived from a single polymorphic type. For example the $Set$ type constructor builds types in the form $Set[T]$ for some given element type $T$, thus we can monomorphize $Set$ using different element types, such as $Set[Int]$, $Set[Float]$, and $Set[String]$.

For the remainder of this section, we will assume our system supports 5 ground types. These types are inherently monomorphic.
$$
G = \{Boolean, Int, Float, Char, String\}
$$
The number of additional types produced by monomorphizing a type constructor with $n$ input types and using $|G|$ ground types is $|G|^n$. For example, monomorphizing $Set$ produces 5 types and monomorphizing a unary function $A \rightarrow R$ produces 25 types. After monomorphizing all type constructors in section~\ref{sec:type-constructors} with the ground types of $G$ there would be an additional 835 types (not including the 5 ground types). This does not cover any nesting of data structures, function types which operate on data structures, or higher order functions. For full coverage of these types, each type constructor would need to be monomorphized using the additional 835 types, resulting in excess of 400 billion types in total.

The following sections will detail how various contemporary GP systems utilize monomorphized types and in particular discuss how the combinatoric explosion limits the viability of the system as the number of complex types grows.

\subsubsection{PushGP}

PushGP evolves programs in the Push language which represents programs as nested sequences of literals and instructions~\cite{Spector:2005:push3}. Program execution utilizes one stack per data type, and instructions take arguments from specific stacks and return values to specific stacks. The number of stacks grows according to the number of monomorphized types. Each stack requires a set of related instructions for manipulating its elements. Thus the combinatoric explosion of types results in an even more severe explosion in instructions, which would result in program search spaces that are too large to search efficiently.

To mitigate this issue, PushGP systems often require external specification of a smaller set of types and instructions which are curated for the specific problem domain. PushGP can solve problems with around 5 types and over 100 instructions~\cite{Helmuth:2015:BenchmarkSuite, Helmuth:2021:GECCO:PSB2}. However, we hypothesize its performance will suffer if required to handle even the 25 types required just to monomorphize $Map$ with the 5 types in $G$, let alone if it were asked to handle arbitrary single-layer nesting of the type constructors in Section~\ref{sec:type-constructors}. Producing polymorphic Push programs where the input values could be any type is impossible, as is allowing arbitrary nesting of type constructors.


\subsubsection{Grammatical Evolution (GE) and Grammar Guided Genetic Programming (G3P)}

GE uses context-free grammars in Backus-Naur form to translate sequences of codons (typically integers) into an abstract syntax trees (AST) by resolving each codon to a particular derivation rule of the grammar~\cite{Ryan:1998:GE}. G3P similarly uses grammar rules to generate and evolve program trees\cite{Forstenlechner:2017:G3P, Forstenlechner:2018:G3P-extention}. In order to ensure type safety in the ASTs the grammar uses a separate derivation rule for expressions of each data type. 

Up to this point, grammar-based approaches have monomorphized each type constructor to pair it with any necessary ground types, such as creating separate grammar rules for $Vector[Int]$, $Vector[String]$, etc., depending on the problem~\cite{Forstenlechner:2017:G3P,Forstenlechner:2018:G3P-extention}. By doing so, the type safety of the system is entirely determined by the grammar rules, instead of using a system designed for polymorphism such as the HM type system. Monomorphizing results in a combinatoric explosion of grammar rules and types, as discussed above, when used to monomorphize all possible types, especially nested data structures.

One might consider creating dedicated derivation rules for each polymorphic type; however, this presents some challenges. Suppose there were grammar rules for a polymorphic $Vector$, such that it could hold elements of different types. Presumably, there would be rules to retrieve an element from an index of the vector. However, the grammar could not know what data type this element is, so it could not have rules that allow it to use such an element with other operators that require specific types, even as simple as allowing two elements to be added together. Context-free grammars (CFG) are unable to ensure type safety of polymorphic expressions because they require type information that is dependant on the context of the call site. This is why the principal use of CFG in practice is to verify program syntax, rather than with semantic verification such as type checking. Thus we see no possible approach to supporting polymorphic data types in grammar-based GP approaches.

\subsection{Polymorphism in CBGP}

Code-building GP is described in detail in~\cite{Pantridge:2022:GECCO:fCBGP} and summarized in section~\ref{sec:CBGP}. The stack-based compilation process composes ASTs by leveraging the type unification algorithm from the HM type system. This means that CBGP has full support for all type constructors, including arbitrary amounts of nested structures. In addition, the function set does not experience a combinatoric explosion because there is only 1 (polymorphic) function per logical transformation over polymorphic types. For example, CBGP includes 1 function for reversing a vector rather than 1 reverse function per concrete vector type.

\subsection{Strongly Typed Genetic Programming (STGP)}

Multiple STGP systems that support at least a limited form of polymorphism have been proposed. Perhaps the most comparable to this work is PolyGP which evolves typed expression trees using subtree mutation and crossover~\cite{Yu:1997:PolyGP}. PolyGP also used the unification algorithm of the HM type system to guide the construction of type-safe programs, however it does not support expressions for function abstraction (aka anonymous functions) or \verb|let| bindings for creating reusable local variables. In contrast with CBGP, PolyGP uses recursion instead of HOF for traversal of lists. PolyGP, and other STGP systems, typically only included the $Vector$ type constructor. It has also been observed that many tree-based GP systems suffer from program bloat that can hinder search performance~\cite{Poli:2008:field-guide-to-gp}.

\section{Benchmark Problems}
\label{sec:problems}

Existing general program synthesis benchmark suites were designed to be used with systems that monomorphize a small number of type constructors, primarily vectors of the ground types in $G$~\cite{Helmuth:2015:BenchmarkSuite, Helmuth:2021:GECCO:PSB2, Helmuth:2022:GPEM}. Inductive program synthesis systems outside of GP have almost exclusively been tested on benchmark problems that require a small set of data types from a domain-specific language that contains a limited set of functions~\cite{Balog:2016:DeepCoder}.

Since we want to exhibit CBGP's ability to handle a large set of polymorphic types and other related capabilities, we decided to design our own suite of benchmark problems. These problems are designed to be non-trivial, in that they require more than a single function call, but are otherwise not intended to push the envelope of difficulty in general program synthesis. Instead, we designed a set of benchmarks that exhibit the following properties that no prior general program synthesis GP system has tackled concurrently:
\begin{itemize}
    \item Problems that use type constructors for data types not previously used in GP, specifically $Set$, $Map$, and $Tuple$ data structures.

    \item Problems that require nested data structures, such as\\ $Vector[Vector[Int]]$ and $Set[Tuple[Int]]$.

    \item Problems whose solutions are higher-order functions that take a function as one of their arguments.

    \item Problems whose solutions are themselves polymorphic functions that are required to run on arguments of different data types. For example, the min-key problem takes as its argument a map where the keys can be any single type, the values are integers, and the program must return the key with the minimum value. Thus the type of this problem is
    \[ min\text{-}key: \forall \alpha. Map[\alpha,Int] \rightarrow \alpha \]
\end{itemize}

\input{problem_table.tex}

Table~\ref{table:problems} lists the 17 problems we use in our experiments, along with the data structures they require and other properties. We have included at least 4 problems that exhibit each data structure and each property. Below we give English-language descriptions of each problem:

\begin{enumerate}[itemsep=5pt,label=\arabic*.]
\item	\textbf{area-of-rectangle}	--	Given two tuples of floats representing the upper-right and lower-left Cartesian coordinates of a rectangle, find the area of the rectangle.
\item	\textbf{centimeters-to-meters}:	--	Given a length in centimeters, return a tuple of (meters, centimeters) that corresponds to the same length.
\item	\textbf{count-true} --	Given a vector of any type $T$ and a predicate function of type $(T) \rightarrow Boolean$, return the count of the number of elements in the vector that make the predicate true.
\item	\textbf{filter-bounds}	--	Given a set of elements that are all of the same comparable type $T$ , and two instances of type $T$ representing a lower and upper bound, filter the set to only include the elements that fall between two bounds (inclusively). This is the only problem not new in this work; it was previously studied in~\cite{Pantridge:2020:GECCO:CBGP}, which used vectors as the data structures instead of sets.

\item	\textbf{first-index-of-true}	--	Given a vector of type $T$ and a predicate function of type $(T) \rightarrow Boolean$, return the first index in the vector where the predicate is true.
\item	\textbf{get-vals-of-key}	--	Given a vector of maps\\ $Vector[Map[String, Int]]$ and a key that exists in each map, return a vector of the values of that key from all of the maps.
\item	\textbf{max-applied-fn}	--	Given an integer $0 < X < 50$ and a function of type $(Int) \rightarrow Int$, return the integer in the range $[0, X)$ that results in the largest value when the function is applied to it.
\item	\textbf{min-key}	--	Given a $\forall \alpha . Map[\alpha, Int]$ where the keys are of some type $\alpha$, return the key with the minimum value.
\item	\textbf{set-cartesian-product}	--	Given two sets of integers, return their Cartesian product, which will be a set of tuples $Set[Tuple[Int,Int]]$.
\item	\textbf{set-symmetric-difference}	--	Given two sets of integers, return their symmetric difference, which will be a set of integers.
\item	\textbf{sets-with-element}	--	Given a set of sets of integers and a target integer, filter the set to only contain sets that contain the target.
\item	\textbf{simple-encryption}	--	Given a string and a function of type \\ $(Char) \rightarrow Char$, apply the function to each character to encrypt the string.
\item	\textbf{sum-2-vals}	--	Given a $Map[String,Int]$ and two strings that are keys of the map, look up the values associated with those keys in the map and return their sum.
\item	\textbf{sum-2-vals-polymorphic}	--	Given a $\forall \alpha . Map[\alpha, Int]$ and two instances of $\alpha$ that are keys of the map, look up the values associated with those keys in the map and return their sum.
\item	\textbf{sum-2D	}--	Given 2-dimensional vector of integers\\ $Vector[Vector[Int]]$, return the sum of all elements.
\item	\textbf{sum-vector-vals}	--	Given a $Map[String,Int]$ and vector of strings that are keys of the map, look up the values associated with those keys in the map and return their sum.
\item	\textbf{time-sheet}	--	Given a vector of tuples\\ $Vector[Tuple[String,Int]]$, where the strings represent\\ names and the integers represent hours worked, and given a specific name, return the sum of the hours associated with that name.
\end{enumerate}

We define problem-specific lists of constants and ephemeral random constants that are used as genes in the CBGP genomes. Additionally, each problem automatically generates training and test data tailored to the problem. The error function, which determines how close a program's output is to the expected output, is based on the type of the output as follows:

\begin{itemize}
    \item $Int$ or $Float$ -- absolute difference.

    \item $String$ -- Levenshtein string edit distance.

    \item $Set[T]$ -- Jaccard distance of sets.

    \item $Tuple[Int,Int]$ -- absolute difference on each of the tuple components.

    \item $T$ of any type -- This only occurs when the problem's output is polymorphic. Here, we simply give an error of 0 if the program's output is correct, and of 1 otherwise.
\end{itemize}

\section{Experimental Methods}
\label{sec:methods}

\begin{table}[t]
    \rowcolors{2}{gray!15}{white}
    \centering
    \begin{tabular}{l l}
      \toprule
      \textbf{Hyperparameter} & \textbf{Value} \\ \midrule
       Population Size & 1000 \\
       Max Generations & 300 \\
       Parent Selection & Lexicase Selection~\cite{Helmuth:2015:UncompromisingLexicase} \\
       Variation & UMAD~\cite{Helmuth:2018:GECCO:UMAD} \\
       Mutation Rate & 0.1 \\
       Initial Genome Sizes & [50, 250] \\
       Number of Training Cases & 200 \\
       Number of Unseen Test Cases & 2000 \\
       Runs per Problem & 100 \\
       \bottomrule
    \end{tabular}
    \caption{The evolutionary hyperparameters used for all CBGP runs in our experiments.}
    \label{table:hyperparameters}
\end{table}

We only conduct experiments using CBGP; we do not provide any comparison experiments with other systems, since, as discussed in Section~\ref{sec:polymorphism-gp}, we know of no other modern GP system that can handle the polymorphic and data structure requirements for our benchmark problems.
The CBGP hyperparameters for our experiments are given in Table~\ref{table:hyperparameters}. Each parent is selected using lexicase selection~\cite{Helmuth:2015:UncompromisingLexicase}. Linear parent genomes are mutated to create children using uniform mutation with additions and deletions (UMAD), a technique first used on linear Push genomes but applicable to any variable length linear genome~\cite{Helmuth:2018:GECCO:UMAD}.

In our implementations of CBGP, the abstract syntax trees produced by genome compilation are evaluated into functions native to the host language Clojure. Being a Lisp dialect, Clojure was chosen because it is trivial to generate program code from data structures. Also Clojure runs on the Java Virtual Machine, which allows evolved programs to be pre-compiled to Java bytecode for superior efficiency compared to other GP program execution models. From anecdotal experience, running a single generation in CBGP takes about 10 seconds on a relatively modern laptop workstation, where generations of similar runs in other Clojure-based GP systems take many minutes.

We generate 200 random training cases for each run that are used to evaluate each program. If a program is found that passes all of the training cases, evolution is halted, and the program is tested on the 2000 unseen test cases. If it also passes those cases, the run is considered successful; if it fails one or more test cases, or if no program is found that passes all of the training cases within 300 generations, the run is considered failed.

In order to determine which functions are included in the \textit{genetic source} for each problem, we adopt the approach taken by PushGP and G3P for general program synthesis. We construct the set of ground types that are relevant to each problem, and include any functions that make use of at least one of those data types.\footnote{Note PushGP has typically taken the opposite approach of only including functions where \emph{all} related types are included in the set specified by the genetic user~\cite{Helmuth:2020:ALife:source}} Additionally, we always use all polymorphic functions and type constructors, so that data structures and anonymous functions can be built no matter what ground types are used. This ensures that CBGP programs could create any possible data structures built out of the ground types.

\begin{table}[t]
\centering
\rowcolors{3}{gray!15}{white}
\begin{tabular}{lrl}
\toprule
\textbf{Type}        & \textbf{Count} & \textbf{Examples}                        \\
\midrule
Polymorphic & 101            & max, first, set-union, map-get  \\
Integer     & 24             & int-add, range, index-of-char   \\
Float       & 15             & double-mult, sin, zero-double?  \\
Boolean     & 12             & and, empty-str?, zero-int?      \\
Char        & 20             & last-str, index-of-char, digit? \\
String      & 31             & join, subs, char-in?            \\
\textbf{Total}       & \textbf{166}            &                                 \\
\bottomrule
\end{tabular}
\caption{For polymorphic and each ground type in CBGP, we list the number of functions included when specifying to include functions of that type. We also give a few examples of each type of function.}
\label{table:function-types}
\end{table}

Note that 101 out of the total 166 functions in our CBGP implementation are polymorphic (in that at least one of their parameters can be any type or is a type constructor that contains any type), meaning only a small proportion of the functions are added by specifying one or more types. Table~\ref{table:function-types} gives the number of functions added by listing each type; note that the total sums to more than 166, because some functions are added by more than one type.

\section{Results}
\label{sec:results}

\input{results_table.tex}

Table~\ref{table:results} presents the primary results of our experiments with CBGP. For each problem, we report the number of runs out of 100 that find a program that passes all training cases and generalizes to the unseen test set. 

Additionally, we are interested in exploring the number of data types that are produced by some program during evolution. These include ground types, the types of the functions in our function set, and the types of all ASTs constructed during evolution. For each run, we count the number of unique types produced, and present the median across all runs of a problem. We ignore differences in the names of type variables when comparing the uniqueness of a type. Because many constructed types occur extremely rarely,\footnote{In fact, the median frequency of how often a type occurred in a run was 2 for every problem.} we are also curious as to the number of types that occur frequently throughout evolution. In order to tabulate frequently-appearing types, Table~\ref{table:results} also gives the number of types that appear at least 1000 times per run.

Note that type counts are incremented every time a type appears in a program, for every program in every generation. Thus runs that find solutions quickly will have less time to produce esoteric constructed types than runs that last a full 300 generations. We hypothesize that this explains the relatively small number of types for the count-true problem, as well as other problems with high success rates.

\subsection{Example Solution Programs}

\begin{figure}[t]
\begin{verbatim}
(defn filter-bounds
  [input1 input2 input3]
  (set (filter (partial > input3)
               (filter (fn [a-30621229]
                         (> a-30621229 input2))
                       (vec input1)))))

(defn max-applied-fn
  [input1 input2]
  (last (sort-by input2
                 (range input1))))

(defn first-index-of-true
  [input1 input2]
  (index-of (map input2 input1)
            true))

(defn time-sheet
  [input1 input2]
  (reduce +
          (map second
               (get (group-by first input1)
                    input2))))

(defn sum-2D
  [input1]
  (reduce +
          (mapcat reverse input1)))

\end{verbatim}
\caption{A sample of solution Clojure programs evolved by CBGP. Anonymous function argument symbols were generated using a unique natural number prefixed with an \texttt{a-}. Whitespace was adjusted for readability.}
\label{fig:example-solutions}
\end{figure}

Figure~\ref{fig:example-solutions} presents solution programs to some of the problems, chosen to exemplify some of the programming themes discussed in this work.

The solution to filter-bounds creates two new functions, both in different ways. The first \verb|filter| in the function uses \verb|partial| to apply the first argument to \verb|>|, making a function that checks if its argument is less than \verb|input3|. The second \verb|filter| uses an anonymous function that checks whether its argument is greater than \verb|input2|. This combination of filters produces the desired functionality.

The problem max-applied-fn takes a function as its second parameter. To solve this problem, the evolved function correctly produces the range from 0 to \verb|input1|, and then uses \verb|sort-by| on \verb|input2| to arrange those integers based on their output when the function \verb|input2| is applied to them. This puts the correct output at the end of the resulting vector, so \verb|last| returns it.

The solution to first-index-of-true converts the input vector into a vector of Booleans using \verb|map|, and then is able to simply find the index of the first appearance of \verb|true|.

The parameters to the time-sheet problem are of the types\\ $Vector[Tuple[String,Int]]$ and $String$, and the return type is $Int$. Note that problem's type signature does not indicate that a $Map$ would be useful for solving this problem. Even so, the evolved solution uses the \verb|group-by| function to create a map where the keys are strings from the tuples and the values are vectors of the tuples themselves. It can then retrieve the tuples corresponding to the correct name using \verb|get| on this map, instead of a more conventional approach that would use \verb|filter| to reduce the \verb|input1| vector to only include the tuples where the first element was the correct name. The rest of the function simply grabs the second element of each remaining tuple and then sums them.

Finally, the solution program to sum-2D must sum all of the integers in a $Vector[Vector[Int]]$. A simple way to do so is to concatenate all of the internal vectors together and then use \verb|reduce| to sum their contents, which this program does. However, the way in which it concatenates the internal vectors is interesting. The \verb|mapcat| function applies a function (in this case \verb|reverse|) to each element of vector (in this case the the elements are the rows of \verb|input1|), and then concatenates the resulting vectors. In fact, here \verb|mapcat| could be passed the \verb|identity| function, but we did not include one in our function set, because we did not imagine a use case. Fortunately order does not matter when summing numbers and evolution was able to find \verb|reverse| as a suitable alternative.

\section{Discussion}

CBGP was able to solve every problem except for one in at least 1 run out of 100. These results show that CBGP is capable of handling problems with diverse sets of requirements regarding type constructors, nested data structures, higher-order functions, and polymorphism. Programs made use of the three data structures not used in previous program synthesis GP systems, $Map$, $Set$, and $Tuple$.

The number of data types created during the CBGP runs is a few orders of magnitude higher than the number of data types handled by other program synthesis GP systems, which handle at most 5 to 10 monomorphized types. Even the number of data types that occur ``frequently'', of at least 1000 times per run, number over 100 for all but one problem.
We expect that many of these represent common ground types and function types, as well as type constructors that are useful for the problem.
Most of the more unusual types appear only a few times throughout evolution. Such types are likely composed of deeply nested function or data structure types, have little or no evolutionary advantage, and disappear from the population quickly. 


\section{Conclusions}

In this paper we expand the kinds of problems that CBGP can tackle in terms of type constructors and polymorphic functions. We show that CBGP can solve problems exhibiting a variety of properties that no prior GP system has tackled all at once: parametric polymorphism, nested data structures, higher-order functions, and a large, general-purpose function set. Compared to other modern general program synthesis GP systems and other non-GP inductive program synthesis systems, CBGP can handle exponentially more data types and/or functions within one GP run while still maintaining its ability to find solutions.

Our results further show that CBGP does produce a large number of data types during its runs, including many that are produced thousands of times per run. These results show that CBGP is not simply ignoring the huge space of possible types, but builds parts of programs with many types throughout evolution. We do not have data on which types were used most often by solution programs or in their evolutionary lineages, so such a study will remain for future work.

Anecdotally, we have noted at least a few solution programs that make use of partial function application; we have not noted any use of function composition, though we have only examined a small number of the solutions. Future work should examine more closely how these and other methods of utilizing polymorphism contribute to evolving solution programs.

CBGP still relies on its stack-based compilation algorithm to produce programs from linear genomes. Many avenues are available for alternative compilation technique while staying within the Hindley-Milner type system. We think it would be worthwhile to explore how different compilation procedures could result in different sizes, shapes, and behaviors of programs.

CBGP currently has no way of indicating that a function could be applied to \textit{some} data types; instead, it must be applied to one type specifically, or all of them (through parametric polymorphism). In contrast, \textit{ad-hoc polymorphism} refers to the ability to define functions which can only be applied to a discrete set of types, but can arbitraily vary the behavior of the function for each type. For example, this work included separate \verb|int-add| and \verb|float-add| functions that operate on integers and floats respectively, where it could potentially be more evolvable to have a single \verb|add| function that works on both data types (or even a combination of the two). Some functional programming languages, like  Haskell, implement \textit{ad-hoc polymorphism} using \textit{type classes} that define shared properties that specific types can have, such as the ability to be compared or the ability to be treated like numbers. Other languages implement \textit{ad-hoc polymorphism} in the form of function overloading. 

The other common form of polymorphism, ``subtype polymorphism'', is also a natural fit for CBGP. By defining a hierarchy of types, we would gain the ability to define functions that apply generically to other generic types, such as a single \verb|size| function which can accept instances of any collection type ($Vector$, $Set$, and $Map$).

Adding these additional forms of polymorphism to the type system supporting CBGP would dramatically generalize the function set and reduce its cardinality without sacrificing any expressiveness. In turn, the may allow evolution to make movements in the search space that would otherwise be impossible.

\begin{acks}

We would like to thank Lee Spector, Ryan Boldi, and Nicholas Freitag McPhee for discussions that helped shape this work.
\end{acks}

\bibliographystyle{ACM-Reference-Format}
\bibliography{main}

\end{document}

%% file: problem_table.tex
\begin{table}[t]
\centering
\rowcolors{3}{gray!15}{white}
\begin{tabular}{l | lccc}
\toprule
\textbf{Problem}                  &      \textbf{Structures} & \textbf{Nest} & \textbf{HOF} & \textbf{Poly} \\
\midrule
area-of-rectangle        & tuple           &      &     &      \\
centimeters-to-meters    & tuple           &      &     &      \\
count-true               & vector          &      & x   & x    \\
filter-bounds            & set             &      &     & x    \\
first-index-of-true      & vector          &      & x   & x    \\
get-vals-of-key          & vector, map     & x    &     &      \\
max-applied-fn           &                 &      & x   &      \\
min-key                  & map             &      &     & x    \\
set-cartesian-product    & set, tuple      &      &     &      \\
set-symmetric-difference & set             &      &     &      \\
sets-with-element        & set             & x    &     &      \\
simple-encryption        &                 &      & x   &      \\
sum-2-vals               & map             &      &     &      \\
sum-2-vals-polymorphic   & map             &      &     & x    \\
sum-2D                   & vector          & x    &     &      \\
sum-vector-vals          & vector, map     &      &     &      \\
times-heet                & vector          & x    &     &      \\
\bottomrule
\end{tabular}
\caption{Properties of our benchmark problems. The \textbf{Structures} column lists the data structures relevant to the problem. \textbf{Nest} indicates whether the problem requires using nested data structures. \textbf{HOF} indicates whether the problem requires the creation of a higher-order function. \textbf{Poly} indicates a problem whose parameter (and possibly return value) is a polymorphic data type.}
\label{table:problems}
\end{table}

%% file: results_table.tex
\begin{table}[t]
\centering
\rowcolors{3}{gray!15}{white}
\begin{tabular}{l | rrr}
\toprule
\textbf{Problem}                  & \textbf{Succ} & \textbf{Types} & \textbf{Types $\geq$ 1000} \\
\midrule
area-of-rectangle        & 59        & 11795 & 122        \\
centimeters-to-meters    & 92        & 2122  & 123        \\
count-true               & 100       & 668   & 86         \\
filter-bounds            & 13        & 11699 & 138        \\
first-index-of-true      & 100       & 1372  & 130        \\
get-vals-of-key          & 12        & 6432  & 173        \\
max-applied-fn           & 24        & 6247  & 138        \\
min-key                  & 31        & 15941 & 146        \\
set-cartesian-product    & 0         & 7755  & 148        \\
set-symmetric-difference & 50        & 3696  & 126        \\
sets-with-element        & 4         & 6186  & 154        \\
simple-encryption        & 96        & 1936  & 138        \\
sum-2-vals               & 94        & 1630  & 143        \\
sum-2-vals-polymorphic   & 100       & 1772  & 142        \\
sum-2D                   & 100       & 1373  & 127        \\
sum-vector-vals          & 16        & 6630  & 165        \\
time-sheet                & 2         & 9768  & 174        \\
\bottomrule
\end{tabular}
\caption{Experimental results. \textbf{Succ} measures the number of runs out of 100 that successfully find a program that passes all given training and test cases. \textbf{Types} gives the median number of data types produced per CBGP run. \textbf{Types $\geq$ 1000} gives the median number of data types that were produced at least 1000 times per CBGP run.}
\label{table:results}
\end{table}